\documentclass[11pt]{article}
\usepackage[margin=1in,letterpaper]{geometry}
\usepackage{graphicx} %
\usepackage[bf,sf]{titlesec}
\usepackage{setspace}
\usepackage{titling}
\usepackage{natbib}
\usepackage{amsmath}
\usepackage{amssymb}
\usepackage[table,dvipsnames]{xcolor}
\usepackage{algorithm}
\usepackage{comment}
\usepackage{algpseudocodex}
\usepackage{multirow}
\usepackage{booktabs}
\usepackage{makecell}
\usepackage{subcaption}
\usepackage{fancybox}
\usepackage{fancyvrb}
\usepackage{multirow}
\usepackage[colorlinks]{hyperref}

\newcommand{\argmin}{\mathop{\rm argmin}}
\newcommand{\argmax}{\mathop{\rm argmax}}

\newcommand{\dquote}[1]{``#1''}

\usepackage{url}            %
\usepackage{booktabs}       %
\usepackage{multirow}    
\usepackage{amsfonts}       %
\usepackage{nicefrac}       %
\usepackage{microtype}      %
\usepackage{natbib}
\usepackage{enumerate}
\usepackage{hhline}
\usepackage{makecell}
\usepackage{pifont}

\usepackage{graphicx} %
\usepackage{amsmath}
\usepackage{amsthm}
\usepackage{amssymb}
\usepackage{tikz}
\usepackage{xcolor}
\usetikzlibrary{arrows}

\allowdisplaybreaks

\usepackage{mathrsfs}

\usepackage{hyperref}
\usepackage{bm}

\allowdisplaybreaks

\newcommand{\labs}{\left\vert}
\newcommand{\rabs}{\right\vert}

\newcommand{\real}{\mathbb{R}}

\newcommand{\expect}{\mathbb{E}}

\newtheorem{prop}{Proposition}

\usepackage[capitalize,noabbrev]{cleveref}
\crefname{thm}{Theorem}{Theorems}
\crefname{lem}{Lemma}{Lemmas}
\crefname{cor}{Corollary}{Corollaries}
\crefname{prop}{Proposition}{Propositions}
\crefname{asmp}{Assumption}{Assumptions}
\crefname{defn}{Definition}{Definitions}
\crefname{oracle}{Oracle}{Oracles}
\crefname{fact}{Fact}{Facts}
\crefname{conj}{Conjecture}{Conjectures}
\crefname{rem}{Remark}{Remarks}
\crefname{example}{Example}{Examples}
\crefname{condition}{Condition}{Conditions}
\crefname{exercise}{Exercise}{Exercises}
\crefname{algorithm}{Algorithm}{Algorithms}
\crefname{table}{Table}{Tables}
\crefname{figure}{Figure}{Figures}
\crefname{section}{Section}{Sections}
\crefname{subsection}{Section}{Sections}
\crefname{appendix}{Appendix}{Appendices}
\crefname{message}{Message}{Messages}

\renewcommand{\algorithmicrequire}{\textbf{Input:}}
\renewcommand{\algorithmicensure}{\textbf{Output:}}

\definecolor{red}{rgb}{1, 0, 0}

\definecolor{green}{rgb}{0, 1, 0}

\definecolor{blue}{rgb}{0, 0, 1}

\definecolor{orange}{rgb}{1, 0.4, 0.0}

\input{packages/math_commands}

\newcommand{\pref}{\operatorname{pref}}
\newcommand{\prefree}{\operatorname{pref-free}}

\newcommand{\REF}{\operatorname{ref}}
\newcommand{\DPO}{\operatorname{DPO}}
\newcommand{\RF}{\operatorname{RMF-PO}}
\newcommand{\RB}{\operatorname{RMB-PO}}
\newcommand{\RBP}{\operatorname{RMB-PO+}}

\AtBeginDocument{\setlength\abovedisplayskip{4.5pt}}
\AtBeginDocument{\setlength\belowdisplayskip{4.5pt}}

\usepackage{listings}
\hypersetup{
  colorlinks   = true, %
  urlcolor     = blue, %
  linkcolor    = blue, %
  citecolor   = blue %
}

\usepackage{wasysym}

\makeatletter
\renewenvironment{abstract}{%
    \if@twocolumn
      \section*{\abstractname}%
    \else %
      \begin{center}%
        {\sffamily \bfseries \abstractname\vspace{\z@}}%
      \end{center}%
      \quotation
    \fi}
    {\if@twocolumn\else\endquotation\fi}
\makeatother

\definecolor{lightgray}{gray}{0.95} %
\renewcommand{\algorithmicrequire}{\textbf{Input:}}
\renewcommand{\algorithmicensure}{\textbf{Output:}}
\renewcommand{\algorithmicloop}{\textbf{repeat}}

\newcommand{\textbsf}[1]{\textsf{\textbf{#1}}}

\title{\textbsf{Policy Optimization in RLHF: The Impact of \\ Out-of-preference Data}}

\usepackage{authblk}
\author[1,2]{Ziniu Li\thanks{Equal contribution. Author ordering is determined by coin flip. Emails: \texttt{ziniuli@link.cuhk.edu.cn} and \texttt{xut@lamda.nju.edu.cn}}}
\author[3,4,5]{Tian Xu{$^*$}}
\author[3,4,5]{Yang Yu\thanks{Corresponding author. Email: \texttt{yuy@nju.edu.cn}}}
\affil[1]{The Chinese University of Hong Kong, Shenzhen}
\affil[2]{Shenzhen Research Institute of Big Data}
\affil[3]{National Key Laboratory for Novel Software Technology, Nanjing University}
\affil[4]{School of Artificial Intelligence, Nanjing University}
\affil[5]{Polixir.ai}

\begin{document}
\maketitle

\begin{abstract}

Aligning agents with human preferences is important. This paper examines two classes of alignment methods. The first class operates without explicitly learning a reward model from preference data, with Direct Preference Optimization \citep{rafailov2023direct} emerging as a prominent method within this class. The second class involves methods that explicitly learn a reward function and utilize it to optimize policy on prompts-only data, with Proximal Policy Optimization \citep{schulman2017ppo} standing out as a popular choice. Within this class, we investigate a notable approach that leverages a large amount of prompts, extending beyond those present in the preference dataset. Experiments demonstrate that this approach outperforms other methods on synthetic contextual bandits, which serve as mathematical models for alignment. Additionally, we provide an analysis of source errors in these optimization methods and draw connections with other related research areas, such as imitation learning and reinforcement learning. In essence, our research highlights the importance of integrating out-of-preference data, including the policy's responses to prompts from the preference dataset and new prompts, into the policy optimization.\footnote{A short version of this paper is presented at the tiny paper track of the 12th International Conference on Learning Representations (ICLR), 2024. Code is available at \url{https://github.com/liziniu/policy_optimization}.}

\end{abstract}

\section{Introduction}

Developing trustworthy agents requires alignment with human preferences \citep{russell2010artificial}. A standard practice involves providing a human preference dataset for the agent to learn from. According to utility theory \citep{fishburn1979utility}, preference is connected with a certain reward function. Currently, there are two kinds of alignment methods: 
\begin{itemize}
    \item The first class of methods, referred to as the reward-model-free approach in this paper, does not explicitly learn a reward model but directly optimizes the language model from preference annotations. This class includes popular algorithms such as Direct Preference Optimization (DPO) \citep{rafailov2023direct} and Identity Policy Optimization (IPO) \citep{azar2023general}.
    \item  The second class of methods, referred to as the reward-model-based approaches in this paper, is exemplified by the so-called Reinforcement Learning from Human Feedback (RLHF) framework \citep{christiano2017deep, stiennon2020learning, openai2023gpt4} framework, with Proximal Policy Optimization (PPO) \citep{schulman2017ppo} standing out as a popular choice. In particular, this class of methods trains a reward model from the preference data and subsequently optimizes the language model to improve its responses to prompts.
\end{itemize} 

While both approaches are able to improve performance by leveraging preference data, the superiority of one method over the other remains an open question, crucial for driving future advancements. We briefly explain the challenges in determining this superiority. First, we expect that the reward model, learned from preference data, possesses a certain generalization capability (i.e., through fine-tuning a powerful pre-trained neural network). To improve the language model, we require prompt-response pairs (i.e., input-output pairs) and their associated reward values (i.e., the supervision signals). This raises an important question: how should we select these prompt-response pairs? Notably, reward-model-free approaches utilize previously collected prompt-response pairs from the preference dataset, whereas reward-model-based approaches generate new responses to prompts using the language model, discarding the responses in the preference dataset. Which approach is more effective? And how do these data sources impact the generalization performance?

We explore the above questions by analyzing the errors in the language model's optimization under the framework of contextual bandits \citep{banditalgo}, which serve as mathematical models for alignment. This analysis helps us predict the algorithm's behaviors without the need for extensive experiments. Within this context, the language model is framed as a \dquote{policy} in broader terms. We demonstrate that the policy optimization in these methods corresponds to various forms of Monte Carlo approximations for maximizing the \emph{expected} reward. Notably, the inclusion of out-of-preference data, which includes responses to prompts from the preference dataset and responses to new prompts, enhances the accuracy of the Monte Carlo approximation.

To validate the above ideas, we conduct experiments on contextual bandits with linear function approximation and neural function approximation, respectively. One main experiment, where we manually ensure that the policy shares the same good feature representation with the reward model (thus, they have the same representation power), shows that policy optimization with additional out-of-preference data still improves generalization performance. Other experiments also support this claim. Finally, we provide a discussion about this phenomenon with reference to other fields, such as imitation learning \citep{osa2018survey} and reinforcement learning \citep{sutton2018reinforcement}.

\section{Problem Formulation}

We consider the so-called contextual bandits \citep{langford2007epoch, lu2010contextual} formulation, which serves mathematical models for alignment. Let $s$ and $a$ be the state and action, respectively. We aim to obtain a decision policy $\pi$ that acts optimally in terms of reward maximization:
\begin{align}   \label{eq:opt}
    \pi^{\star}_{r} \lar \argmax_{\pi} \expect_{s \sim \rho(\cdot)} \expect_{a \sim \pi(\cdot|s)} \ls r(s, a) \rs,
\end{align}
where the symbol $\rho$ denotes the state distribution, and $r$ is the ground truth reward function. We omit the subscript $r$ in $\pi^{\star}_r$ when the context is clear. For language models, the term \dquote{states} refers to prompts, while \dquote{actions} denote responses. The language model functions as the decision-making policy. It is worth noting that terminologies may be used interchangeably.

In the context of alignment, the difficulty is that the reward function is unknown but only preferences over two actions are observed. Typically, the Bradley-Terry assumption \citep{bradley1952rank} is used:
\begin{align*}
    \sP \lp a  > a^{\prime} | s \rp = \frac{\exp (r(s, a))}{\exp(r(s, a)) + \exp(r(s, a^{\prime}))},
\end{align*}
where the symbol $a > a^{\prime}$ means that $a$ is more preferred compared with $a^{\prime}$. Given a preference dataset $D_{\pref} = \{(s_i, a_i, a_i^{\prime}) \}_{i=1}^{n}$, where $a_i > a_i^{\prime}$ is assumed without loss of generality, the reward learning objective, derived via maximum likelihood estimation, is 
\begin{align}   \label{eq:learn_reward}
    \widehat{r} \lar \argmax_{r} \sum_{i=1}^{n}  \log \left( \sigma \left( r(s_i, a_i) - r(s_i,a_i^{\prime}) \right) \right), \quad   (s_i, a_i, a_i^{\prime})  \sim D_{\pref}.
\end{align}
where $\sigma(\cdot)$ is the sigmoid function. This objective encourages the reward function to give a high score for the positively preferred data $(s, a)$ and a low score for the negative preferred data $(s, a^{\prime})$.

\begin{figure}[t]
    \centering
    \begin{subfigure}{.75\linewidth}
      \centering
      \includegraphics[width=0.9\linewidth]{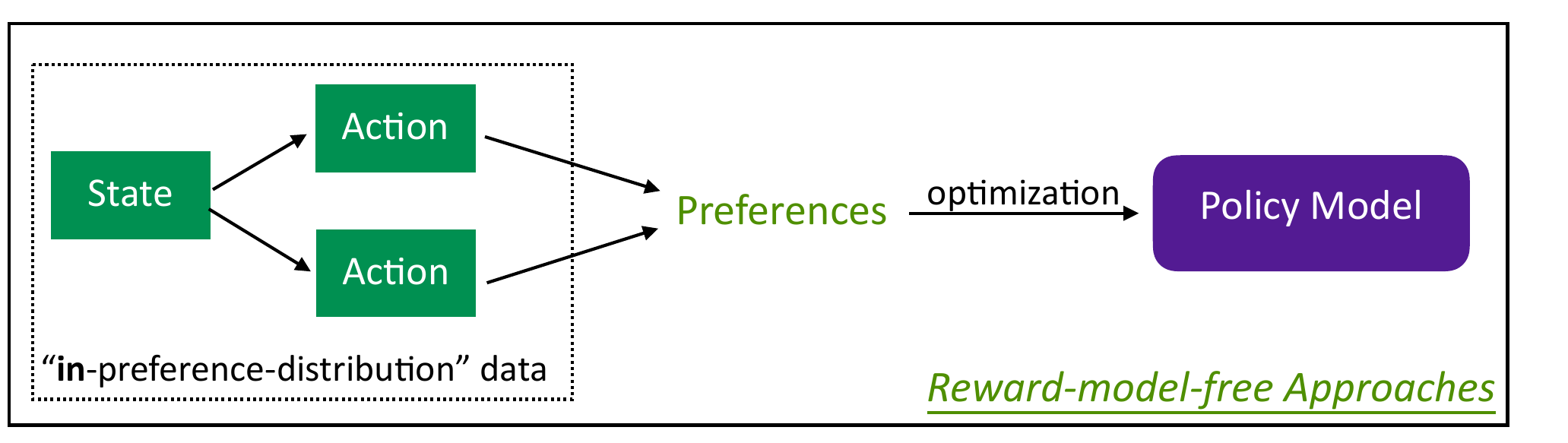}  
      \caption{Illustration for reward-model-free approaches.}
    \end{subfigure}
    \hfill
    \begin{subfigure}{.74\linewidth}
      \centering
      \includegraphics[width=0.9\linewidth]{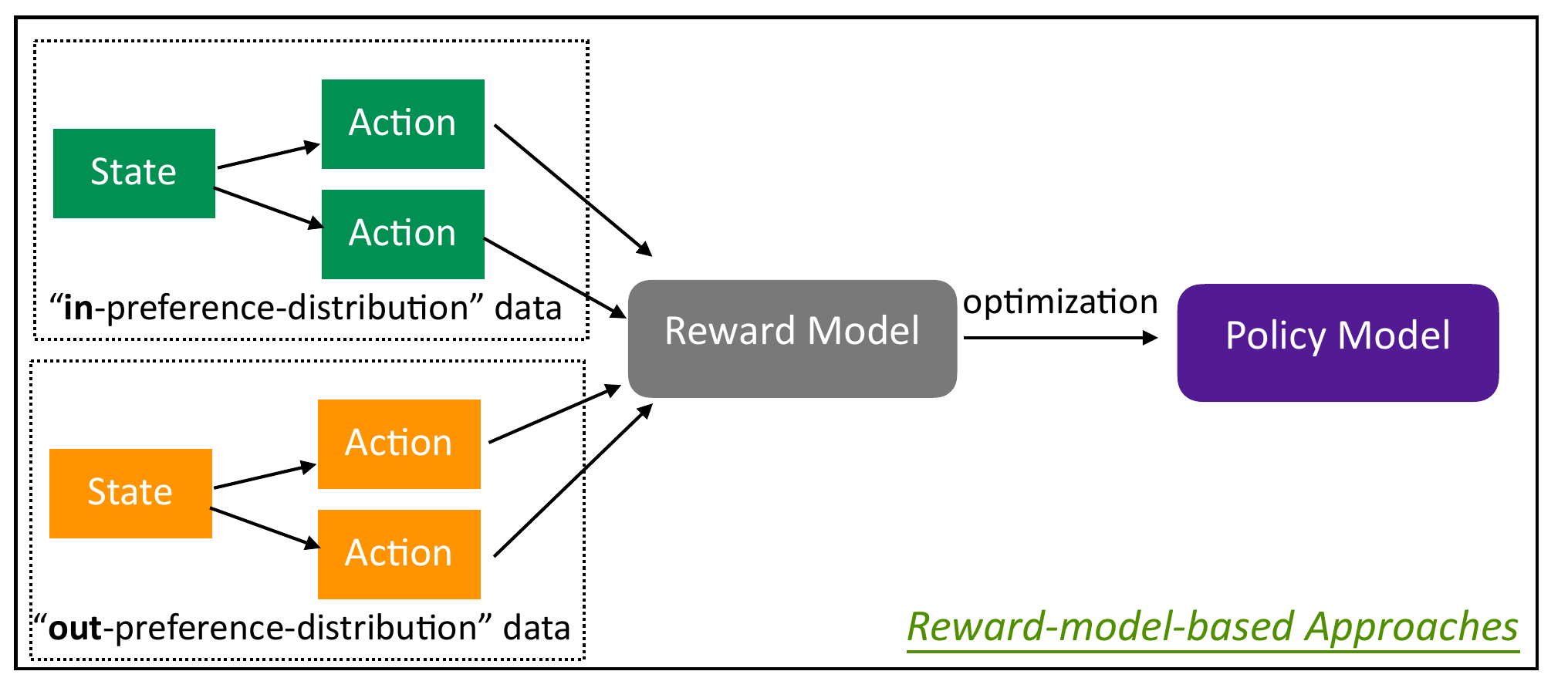} 
      \caption{Illustration for reward-model-based approaches.}
    \end{subfigure}
    \hfill 
    \caption{Illustration for policy optimization methods. For reward-model-based approaches, the reward model learning procedure is not plotted for ease of presentation.}
    \label{fig:main_freamwork} 
\end{figure}

Let $\pi_{\REF}$ be a reference policy model and $\beta > 0$ be a hyper-parameter. 
Ideally, we may want to optimize the policy with this recovered reward function in population:
\begin{align*}
   \pi^{\star}_{\widehat{r}} \lar \argmax_{\pi} \expect_{s \sim \rho(\cdot)} \lb \expect_{a \sim \pi(\cdot|s)} \ls \widehat{r}(s, a)\rs - \beta \KL (\pi(\cdot|s),\pi_{\REF}(\cdot|s)) \rb.
\end{align*}
Here, the Kullback–Leibler (KL) penalty aims to mitigate the reward hacking and over-optimization issue \citep{gao2023scaling}. We remark that, in practice, we do not know the distribution $\rho(\cdot)$ and typically employ Monte Carlo approximations. That is, we use finite samples to approximate the population distribution and its expectation. There are two kinds of practical approaches, which we elaborate on below. Please also see \cref{fig:main_freamwork}.

\textbf{Reward-model-free Optimization Approaches:} One direct idea is to use state-action pairs from the preference dataset for policy optimization:
\begin{align} \label{eq:reward_free_opt}
   \widehat{\pi}_{\RF} \lar \argmax_{\pi} \sum_{i=1}^{n} \sum_{a \in \{a_i, a_i^{\prime} \}} \widehat{r} (s_i, a) - \beta \KL \lp \pi(a|s_i), \pi_{\REF}(a|s_i) \rp, \quad (s_i, a_i, a_i^{\prime}) \sim D_{\pref}.
\end{align}
This approach approximate the state and action distributions by finite samples from the preference dataset. By using tools from KL-regularized optimization (see e.g., \citep{vieillard2020leverage}),  \citet{rafailov2023direct} showed that procedures in \cref{eq:learn_reward} and \cref{eq:reward_free_opt} could be integrated into a single objective:
\begin{align}   \label{eq:dpo}
    \widehat{\pi}_{\DPO} \lar \argmax_{\pi} \sum_{i=1}^{n} \log \sigma \lp \beta  \log \frac{\pi(a_i|s_i)}{\pi_{\REF}(a_i|s_i)} - \beta \log \frac{\pi(a_i^{\prime} |s_i)}{\pi_{\REF}(a_i^{\prime}|s_i)} \rp, \quad (s_i, a_i, a_i^{\prime}) \sim D_{\pref}.
\end{align}
The resultant algorithm is named Direct Preference Optimization (DPO). Since this approach does not require explicitly training a reward model, it is considered as reward-model-free optimization.

\textbf{Reward-model-based Optimization Approaches:} 
The vanilla Reward-Model-Based Policy Optimization (RMB-PO) approach also leverages states from the preference dataset but samples actions from the policy model:
\begin{align}     \label{eq:learn_policy_reward}
    \widehat{\pi}_{\RB} \lar \argmax_{\pi} \sum_{i=1}^{n} \expect_{a \sim \pi(\cdot|s_i)}\ls  \widehat{r}(s_i, a) \rs - \beta \KL (\pi(\cdot|s_i),\pi_{\REF}(\cdot|s_i)), \quad  s_i \sim D_{\pref}.
\end{align}
We consider the exact action expectation $\expect_{a \sim \pi(\cdot|s)}[\widehat{r}(s, a)]$ in the above formulation, and this expectation can be approximated by sampling multiple actions. This approximation error can be mitigated by computational power, and we do not consider this error in this paper.

In \citep{ouyang2022training, touvron2023llama}, a variant of RMB-PO, refereed to as RMB-PO+ in this paper, further leverages a new, \emph{preference-free} dataset $D_{\prefree} = \{ s_j \}_{j=1}^{m}$:
\begin{align}   \label{eq:learn_policy_reward_plus}
    \widehat{\pi}_{\RBP} \lar \argmax_{\pi} \sum_{j=1}^{m}  \expect_{a \sim \pi(\cdot|s_j)}\ls  \widehat{r}(s_j, a) \rs - \beta \KL (\pi(\cdot|s_j),\pi_{\REF}(\cdot|s_j)), \,\,  s_j  \sim D_{\prefree}.
\end{align}
Note that the dataset $D_{\prefree}$ is cheap to obtain and  usually $m \geq n$ \citep{ouyang2022training}. One particular example of such data in language model's application is the \texttt{lmsys-chat-1m} dataset \citep{zheng2023lmsys}, which has 1 million prompts from real users without preference annotations.

We note that there is no single learning objective for reward-model-based approaches. This is because the technique in \citep{rafailov2023direct} requires that the reward and policy learning objectives have the same training distribution, a condition that is not met for reward-model-based approaches. In practice, policy optimization in \cref{eq:learn_policy_reward} and \cref{eq:learn_policy_reward_plus} can be conducted by policy gradient methods such as PPO \citep{schulman2017ppo} and ReMax \citep{li2023remax}.

\subsection{Theoretical Analysis}

In this section, we present a preliminary analysis of errors in the optimization methods. At a high level, we identify three types of errors:
\begin{itemize}
    \item[1)] the reward evaluation error \( |\widehat{r}(s, a) - r(s, a)| \);
    \item[2)] the estimation error when using finite samples to calculate the expectation \( \mathbb{E}_{a \sim \pi(\cdot|s)}[\cdot] \);
    \item[3)] the estimation error when using finite samples to calculate the expectation \( \mathbb{E}_{s \sim \rho(\cdot)}[\cdot] \).
\end{itemize}
The first error primarily results from finite preference data and diminishes to zero as the preference data size increases indefinitely. This error exists in all optimization methods. Compared with DPO, RMB-PO aims to mitigate the second error, while RMB-PO+ further reduces the third error. We note that RMB-PO and RMB-PO+ do not increase the sample complexity of preference data but only incur additional computational steps. We present the error bound analysis below.

\begin{prop}    \label{prop:main}
Define the reward evaluation error $\varepsilon_{r} = \sup_{(s, a)} |\widehat{r}(s, a) - r(s, a)|$, the state distribution estimation error $\varepsilon_s = \sup_{\pi}\sup_{r: \gS \times \gA \rar [0, 1]} \expect_{s}\expect_{a \sim \pi(\cdot|s)}[r(s, a)] - \widehat{\expect}_{s}\expect_{a \sim \pi(\cdot|s)}[r(s, a)]$, and the action distribution estimation error $\varepsilon_{a} = \sup_{\pi}\sup_{r: \gS \times \gA \rar [0, 1]} \sup_{s} \expect_{a \sim \pi(\cdot|s)}[r(s, a)] - \widehat{\expect}_{a \sim \pi (\cdot|s)}[r(s, a)]$. Here $\widehat{\expect}_{s}$ and $\widehat{\expect}_{a}$ denote the finite-sample estimations of expectation under the state and action distributions, respectively. Consider $\beta = 0$, then we have 
\begin{align*}
     r(\pi^{\star}_r) - r(\widehat{\pi}_{\RB}) &\leq  2\varepsilon_{r} + 2 \varepsilon_{s}, \\ 
       r(\pi^{\star}_r) - r(\widehat{\pi}_{\RF}) &\leq 2 \varepsilon_{r} + 2 \varepsilon_{s} + 2 \varepsilon_{a},
\end{align*}
where $r(\pi) = \expect_{s}\expect_{a \sim \pi(\cdot|s)}[r(s, a)]$ is the evaluation performance of a policy $\pi$.
\end{prop}

\begin{proof}
We first consider the reward error:
\begin{align}
    &\quad r(\pi^{\star}_r) - r(\widehat{\pi}_{\RB}) \nonumber  \\
    &= \expect_{s}\expect_{a \sim \pi^{\star}_r}[r(s, a) ] - \expect_{s} \expect_{a \sim \widehat{\pi}_{\RB}}[r(s, a)] \nonumber \\
    &= \expect_{s}\expect_{a \sim \pi^{\star}_r}[r(s, a) ]  - \expect_{s}\expect_{a \sim \pi^{\star}_r}[\widehat{r}(s, a) ]  + \expect_{s}\expect_{a \sim \pi^{\star}_r}[\widehat{r}(s, a) ] -  \expect_{s} \expect_{a \sim \widehat{\pi}_{\RB}}[r(s, a)] \nonumber  \\
    &\leq \varepsilon_{r} + \expect_{s}\expect_{a \sim \pi^{\star}_r}[\widehat{r}(s, a) ] - \expect_{s} \expect_{a \sim \widehat{\pi}_{\RB}}[\widehat{r}(s, a)] + \expect_{s} \expect_{a \sim \widehat{\pi}_{\RB}}[\widehat{r}(s, a)] -  \expect_{s} \expect_{a \sim \widehat{\pi}_{\RB}}[r(s, a)] \nonumber  \\
    &\leq 2 \varepsilon_{r}  + \expect_{s}\expect_{a \sim \pi^{\star}_r}[\widehat{r}(s, a) ] - \expect_{s} \expect_{a \sim \widehat{\pi}_{\RB}}[\widehat{r}(s, a)] \label{eq:proof_1}.
\end{align}
Then we consider the state distribution estimation error:
\begin{align}
     &\quad \expect_{s}\expect_{a \sim \pi^{\star}_r}[\widehat{r}(s, a) ] - \expect_{s} \expect_{a \sim \widehat{\pi}_{\RB}}[\widehat{r}(s, a)] \nonumber  \\ 
    &= \expect_{s} \expect_{a \sim \pi^{\star}_r}[\widehat{r}(s, a)] - \widehat{\expect}_{s} \expect_{a \sim \pi^{\star}_r}[\widehat{r}(s, a)] + \widehat{\expect}_{s} \expect_{a \sim \pi^{\star}_r}[\widehat{r}(s, a)] - \expect_{s} \expect_{a \sim \widehat{\pi}_{\RB}}[\widehat{r}(s, a)] \nonumber \\
    &\leq \varepsilon_{s} +  \widehat{\expect}_{s} \expect_{a \sim \pi^{\star}_r}[\widehat{r}(s, a)] - \widehat{\expect}_{s} \expect_{a \sim \widehat{\pi}_{\RB}}[\widehat{r}(s, a)]  + \widehat{\expect}_{s} \expect_{a \sim \widehat{\pi}_{\RB}}[\widehat{r}(s, a)] -  \expect_{s} \expect_{a \sim \widehat{\pi}_{\RB}}[\widehat{r}(s, a)] \nonumber \\
    &\leq \varepsilon_{s} + 0 + \varepsilon_{s}. \label{eq:proof_2}
\end{align}
Combining \eqref{eq:proof_1} and \eqref{eq:proof_2} proves the first result in \cref{prop:main}. For the second result, we may replace $\widehat{\pi}_{\RB}$ with $\widehat{\pi}_{\RF}$ in the above proof to obtain:
\begin{align}  \label{eq:proof_3}
    r(\pi^{\star}_r) - r(\widehat{\pi}_{\RF}) \leq 2 \varepsilon_{r} + 2 \varepsilon_{s} +  \widehat{\expect}_{s}\expect_{a \sim \pi^{\star}_r}[\widehat{r}(s, a)] - \widehat{\expect}_{s} \expect_{a \sim \widehat{\pi}_{\RF}}[\widehat{r}(s, a)].
\end{align}
Then we consider the action distribution estimation error:
\begin{align}
    &\quad  \widehat{\expect}_{s}\expect_{a \sim \pi^{\star}_r}[\widehat{r}(s, a)] - \widehat{\expect}_{s} \expect_{a \sim \widehat{\pi}_{\RF}}[\widehat{r}(s, a)] \nonumber \\
    &=  \widehat{\expect}_{s}\expect_{a \sim \pi^{\star}_r}[\widehat{r}(s, a)] - \widehat{\expect}_{s}\widehat{\expect}_{a \sim {\pi}^{\star}_{r}}[\widehat{r}(s, a)] +  \widehat{\expect}_{s}\widehat{\expect}_{a \sim {\pi}^{\star}_{r}}[\widehat{r}(s, a)] - \widehat{\expect}_{s} \expect_{a \sim \widehat{\pi}_{\RF}}[\widehat{r}(s, a)] \nonumber \\ 
    &\leq \varepsilon_{a} + \widehat{\expect}_{s}\widehat{\expect}_{a \sim {\pi}^{\star}_{r}}[\widehat{r}(s, a)] - \widehat{\expect}_{s} \widehat{\expect}_{a \sim \widehat{\pi}_{\RF}}[\widehat{r}(s, a)] + \widehat{\expect}_{s} \widehat{\expect}_{a \sim \widehat{\pi}_{\RF}}[\widehat{r}(s, a)]  - \widehat{\expect}_{s} \expect_{a \sim \widehat{\pi}_{\RF}}[\widehat{r}(s, a)] \nonumber \\
    &\leq \varepsilon_{a}  + 0 + \varepsilon_{a}  \label{eq:proof_4}
\end{align}
Combining \eqref{eq:proof_3} and \eqref{eq:proof_4} proves the second result in \cref{prop:main}.
\end{proof}
We do not present the analysis for $\widehat{\pi}_{\RBP}$, as its analysis is similar to that for $\widehat{\pi}_{\RB}$. The main difference lies in that the state distribution estimation error for $\widehat{\pi}_{\RBP}$ is generally smaller than that for $\widehat{\pi}_{\RB}$ due to more samples. We note that our analysis is quite basic in the sense that we consider all errors in the supremum norm, and it would be interesting to explore a more tighter analysis with finite sample guarantee; see e.g., \citep{xiong2023gibbs} for recent progress in this direction.

\section{Experiments}

In this section, we conduct numerical experiments to validate the improvement of RMB-PO and RMB-PO+ by better stochastic approximation. All of our experiments are run with 10 different random seeds (2021-2030), and the averaged results are reported\footnote{We exclude the worst and best results to make a robust estimation of the performance.}. Note that we set $\pi_{\text{REF}}$ to be a policy with a uniform action distribution in all experiments and $\beta = 0.01$ for all methods. Besides, we use a policy with a uniform action distribution to collect the preference data.

\subsection{Linear Bandit}

We study a linear bandit task, where we have $ r(s, a) = \phi_r(s, a)^{\top} \theta_r^{\star} $, with $ \phi_r(s, a) \in \mathbb{R}^{d} $ denoting the feature representation and $ \theta_r^{\star} \in \mathbb{R}^{d} $ as the parameter. In this case, the reward learning optimization problem is convex, so we use CVXPY \citep{diamond2016cvxpy} to find the solution $\widehat{r}$. In particular, we use the feature map $\phi_r(s, a)$ and the parameter $\theta_r^{\star}$ as 
\begin{align*}
    \phi_{r}(s, a) = \left(  (a+1) \cdot \cos(s \cdot \pi), \frac{1}{a+1} \cdot \sin(s \cdot \pi) \right)^{\top}, \quad \theta_{r}^{\star} = (1, 2)^{\top}, 
\end{align*}
where \( s \in \mathcal{S} = [0, 1] \) and \( a \in \mathcal{A} = \{0, 1, 2, 3\} \). A uniform distribution over $\gS$ is studied.  For the policy, we consider the parameterization 
\begin{align*}
    \pi(a|s) =  \frac{\exp(\phi_{\pi}(s, a)^{\top} \theta_{\pi})}{\sum_{a^{\prime}} \exp(\phi_{\pi}(s, a^{\prime})^{\top} \theta_{\pi})}, 
\end{align*}
with $\phi_{\pi}(s, a)$ and $\theta_{\pi}$ both in $\real^{2}$. In this case, the policy optimization problem is a non-convex problem, but the gradient domination condition holds \citep{agarwal2021theory}. We use the gradient ascent method with the AdaGrad optimizer \citep{duchi2011adaptive} (a step size of 0.1 is used).

\begin{figure}[htbp]
\begin{minipage}[t]{0.47\linewidth}
\centering
\includegraphics[width=0.9\linewidth]{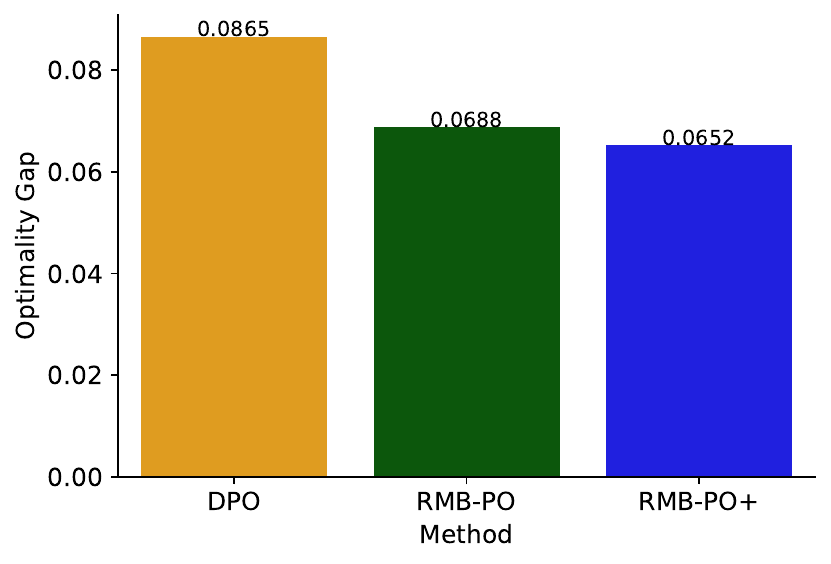}
\caption{  {\small Optimality gap with $\phi_{\pi} = \phi_{r}$.}}
\label{fig:linear_bandit}
\end{minipage}
\hfill
\begin{minipage}[t]{0.47\linewidth}
\centering
\includegraphics[width=0.9\linewidth]{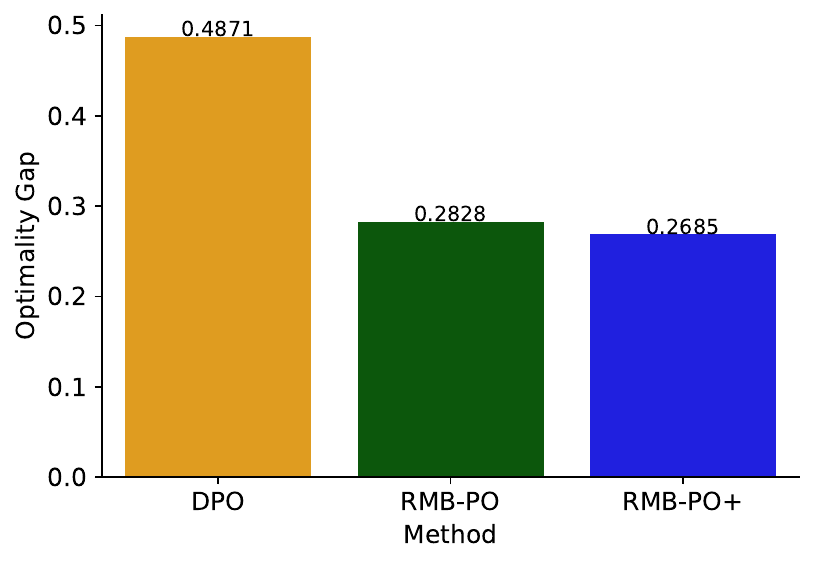}
\caption{ {\small Optimality gap with $\phi_{\pi} \ne \phi_{r}$.}}
\label{fig:linear_bandit_flip}
\end{minipage}
\end{figure}

We examine two scenarios. In the first scenario, there is no feature mismatch between the reward and policy models, i.e., $ \phi_{\pi} = \phi_{r} $. In the second, we use a different feature map for policy:  
\begin{align*}
    \phi_{\pi}(s, a) = \left(  (a+1) \cdot \sin(s \cdot \pi), \frac{1}{a+1} \cdot \cos(s \cdot \pi) \right)^{\top}.
\end{align*}
We believe that in scenarios where $\phi_{\pi} \ne \phi_r$, RMB-PO approaches could exhibit more promising performance than RMF-PO approaches. This is because, in such cases, the policy and reward models may align well by learning from preference data. However, in out-of-preference-distribution scenarios, they may extrapolate and generalize quite differently due to mismatches in representations. Nevertheless, RMB-PO approaches could use out-of-preference-distribution data to mitigate these mismatches and tend to perform well. The case where $\phi_{\pi} \ne \phi_r$ will be revisited in later neural bandit experiments, where the policy model and reward model typically utilize distinct architectures and learn distinct representations.

In our experiments, we set the size of preference data to be $n = 20$ and the size of preference-free data to be $m = 10n$, resulting in training accuracy of the reward model ranging from 60\% to 80\% over 10 experiments. We display the optimality gap $|r(\pi^{\star}) - r(\widehat{\pi})|$ (the smaller, the better) in \cref{fig:linear_bandit} and \cref{fig:linear_bandit_flip}, where $r(\pi)$ is the evaluation performance of a policy $\pi$, i.e., $r(\pi) = \expect_{s \sim \rho(\cdot)} \expect_{a \sim \pi(\cdot|s)}[r(s, a)]$ (in our experiments, we use 5000 sampled states to approximate this expectation).

From \cref{fig:linear_bandit}, we see that even though the policy model is provided with a good feature (e.g., in \cref{fig:linear_bandit}), RMB-PO methods can benefit from out-of-preference data. In the case where $\phi_{\pi} \ne \phi_{r}$ in \cref{fig:linear_bandit_flip}, we find that RMB-PO+ is better than RMB-PO by leveraging additional preference-free data. Thus, we believe it is crucial to learn the optimal action (as inferred by the reward model) on out-of-preference data, even when the two models share the same good feature.

\begin{figure}[H]
    \centering
    \begin{subfigure}{.47\linewidth}
      \centering
      \includegraphics[width=0.95\linewidth]{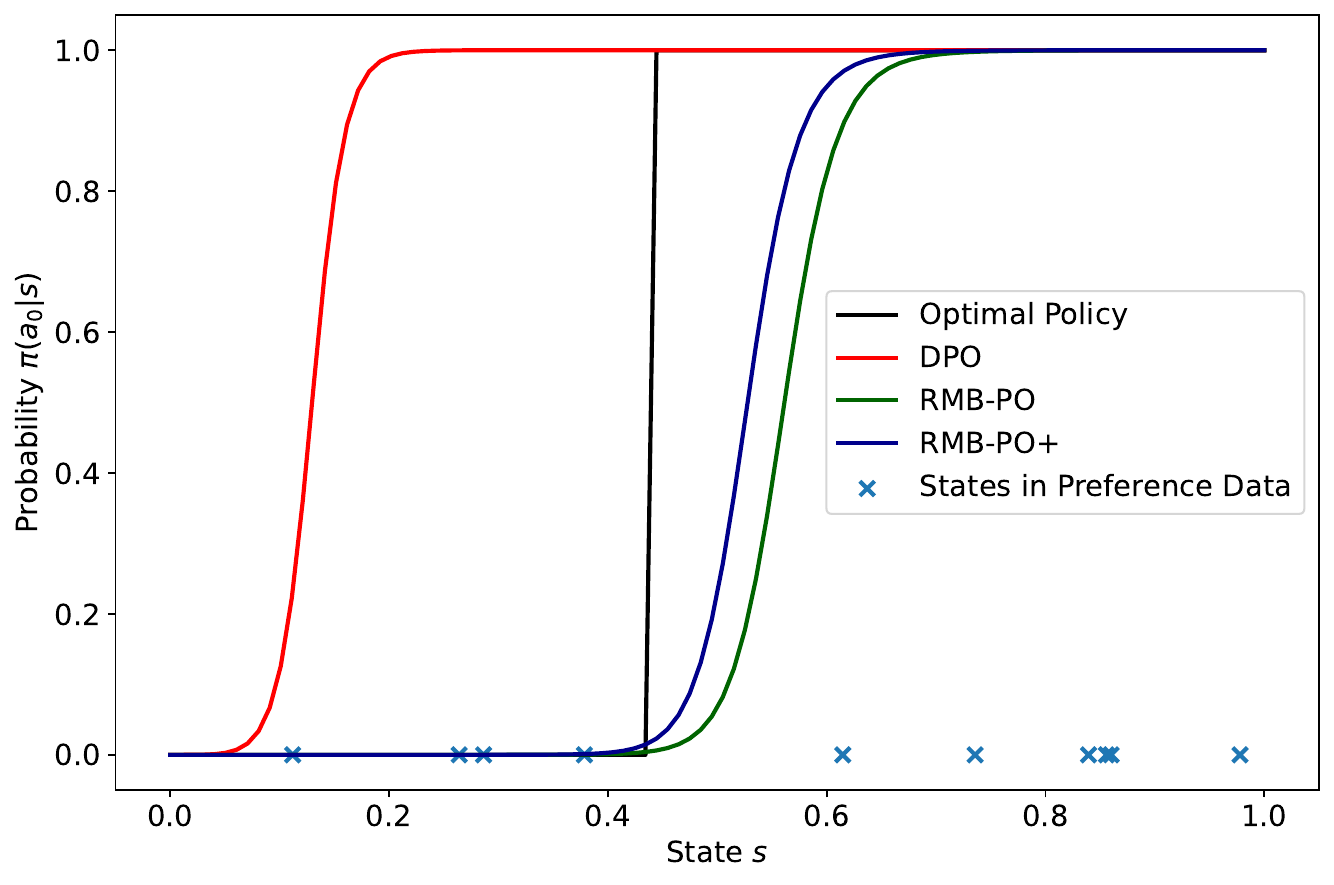}  
      \caption{Action $a_0$.}
    \end{subfigure}
    \hfill
    \begin{subfigure}{.47\linewidth}
      \centering
      \includegraphics[width=0.95\linewidth]{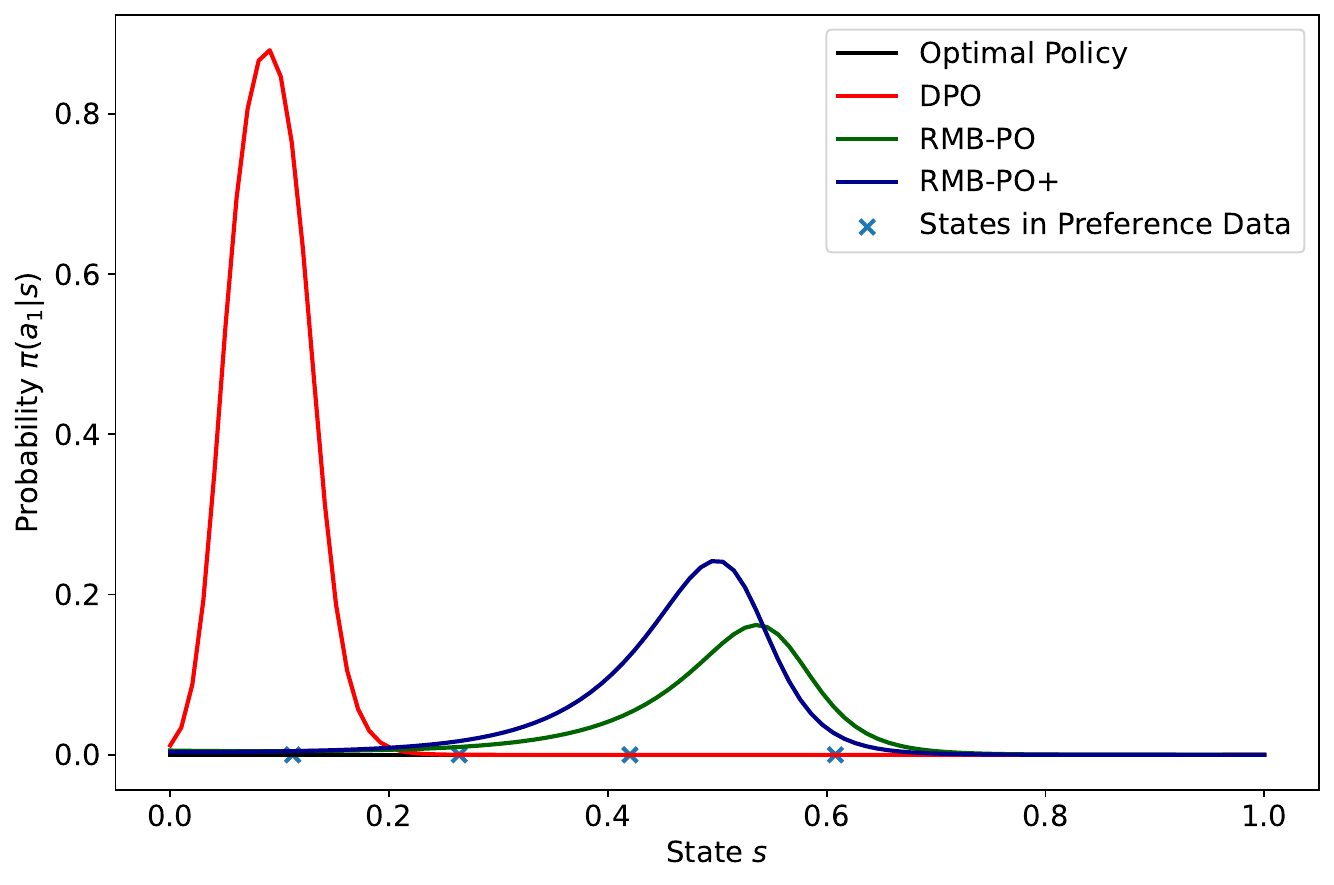} 
      \caption{Action $a_1$.}
    \end{subfigure}
    \hfill 
    \begin{subfigure}{.47\linewidth}
      \centering
      \includegraphics[width=0.95\linewidth]{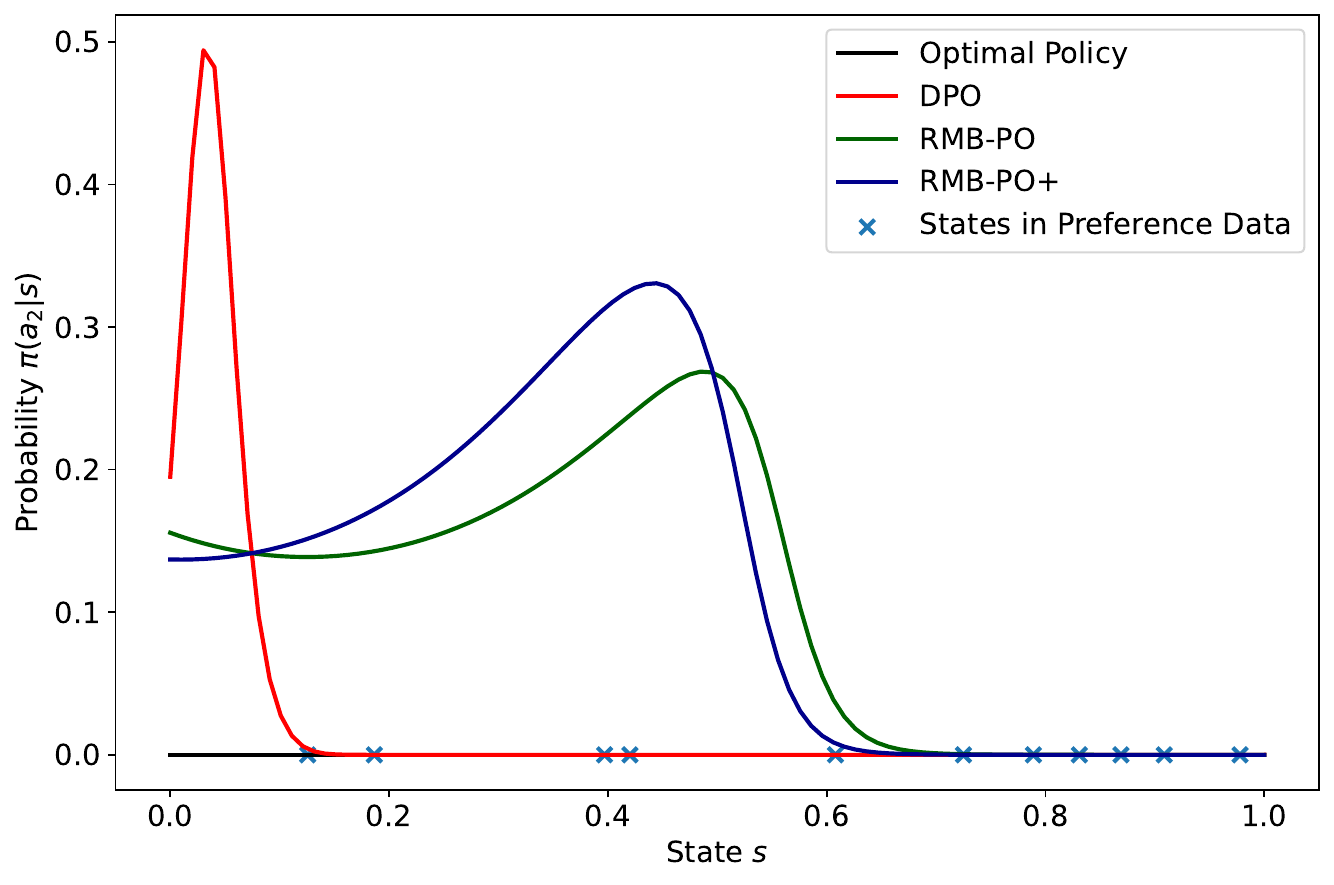} 
      \caption{Action $a_2$.}
    \end{subfigure}
    \hfill 
    \begin{subfigure}{.47\linewidth}
      \centering
      \includegraphics[width=0.95\linewidth]{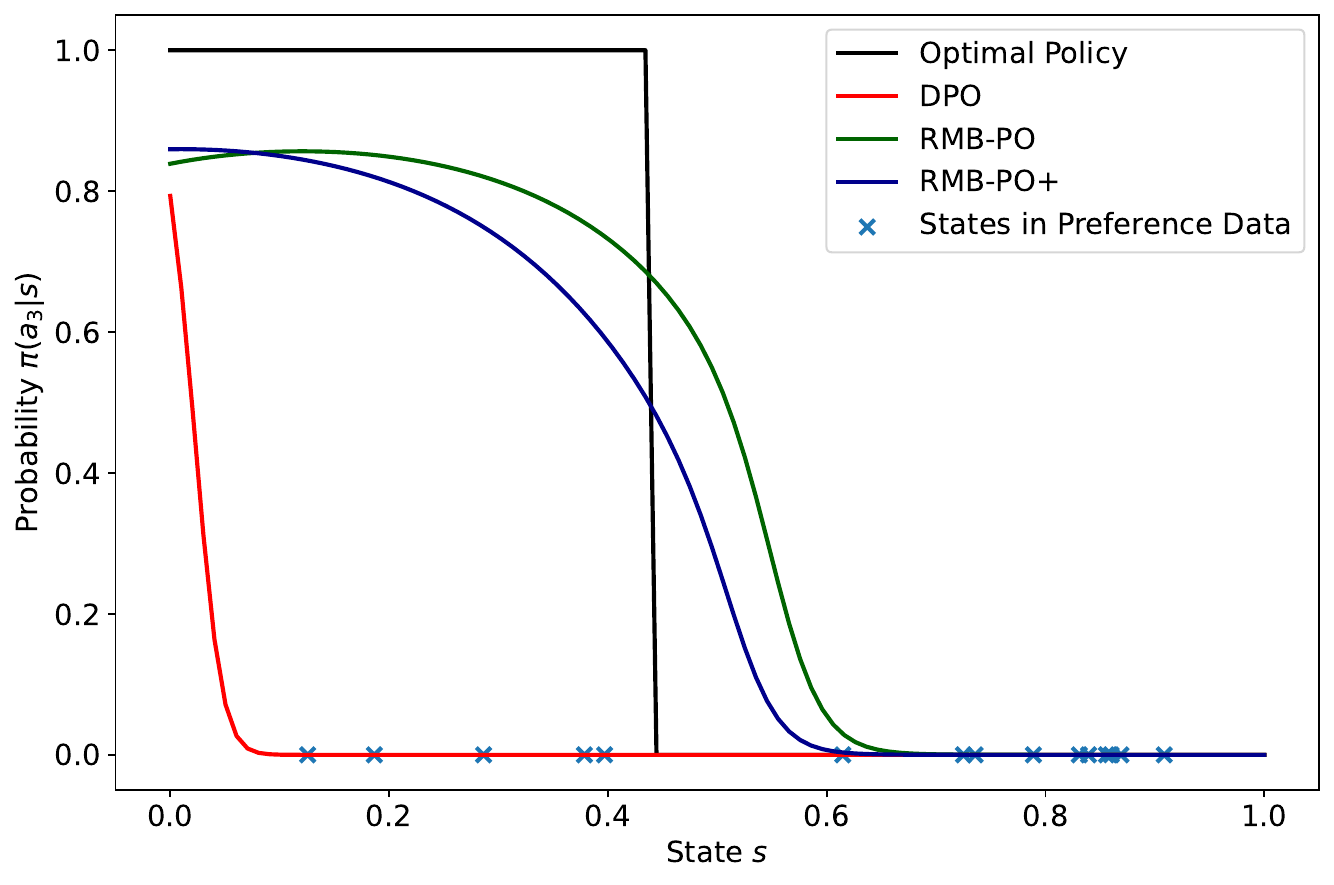} 
      \caption{Action $a_3$.}
    \end{subfigure}
    \hfill 
    \caption{Probabilities of four actions $a_0$, $a_1$, $a_2$ and $a_3$. Results illustrate that RMB-PO(+) methods leverage out-of-preference data to better learn the policy distribution on out-of-distribution states and improve the generalization performance.}
    \label{fig:action probability} 
\end{figure}

To gain a better understanding, we also visualize the learned policy distribution in the $\phi_{\pi} \ne \phi_{r}$ setting; see \cref{fig:action probability}. To observe the training distribution coverage, we plot the states from the preference dataset. Additional states used in RMB-PO+ almost cover the entire state space but are not shown for readability reasons. From the reported curves, we observe that DPO aligns well with the optimal policy in the regions covered by preference data, and RMB-PO(+) methods tend to perform better than DPO in the out-of-distribution regime not covered by the preference data.

\begin{figure}[htbp]
\begin{minipage}[t]{0.47\linewidth}
\centering
\includegraphics[width=0.9\linewidth]{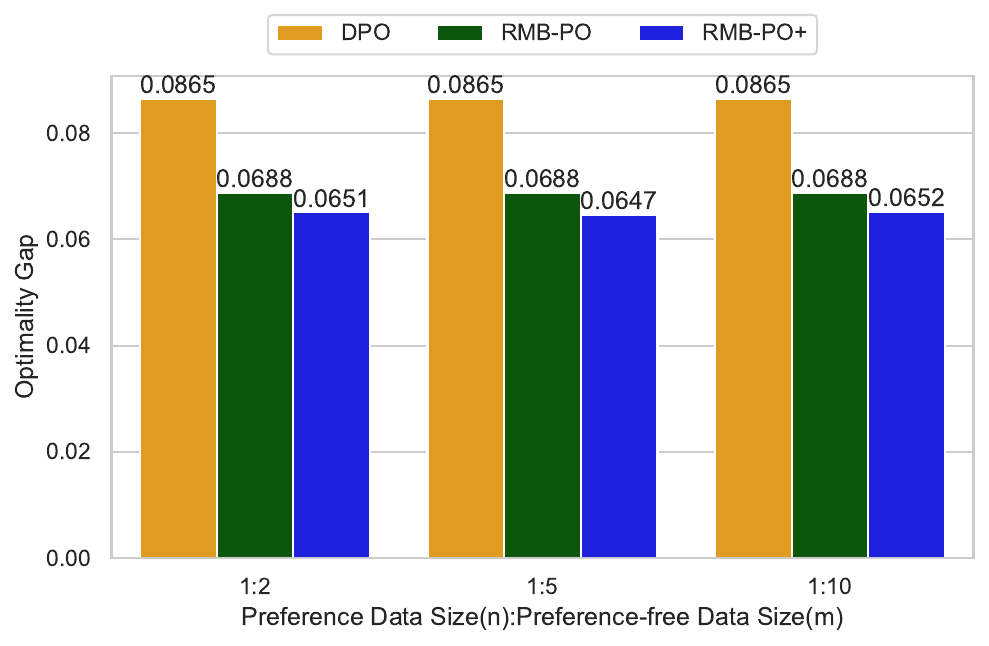}
\caption{  {\small Optimality gap with $\phi_{\pi} = \phi_{r}$.}}
\label{fig:linear_bandit_all}
\end{minipage}
\hfill
\begin{minipage}[t]{0.47\linewidth}
\centering
\includegraphics[width=0.9\linewidth]{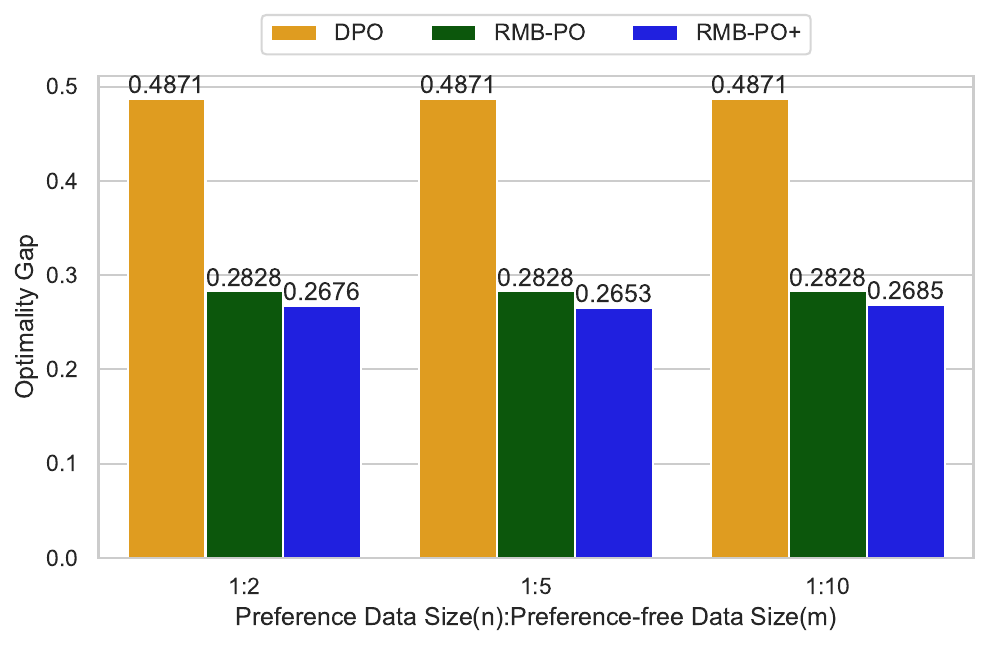}
\caption{ {\small Optimality gap with $\phi_{\pi} \ne \phi_{r}$.}}
\label{fig:linear_bandit_flip_all}
\end{minipage}
\end{figure}

Following the same setup, we provide ablation studies regarding the size of preference-free data used in RMB-PO+. See the results in \cref{fig:linear_bandit_all} and \cref{fig:linear_bandit_flip_all}. We find that the previous conclusions still hold true.

\subsection{Neural Bandit}

In this section, we study a neural bandit problem. Specifically, we study the case where \( r(s, a) = f_{\theta^{\star}_r}(s, a) \), with \( f_{\theta_r^{\star}} \) being a fixed 1-hidden-layer multi-layer perceptron (MLP) neural network, having a hidden size of 64. For reward learning, we use a 2-hidden-layer MLP with a hidden size of 64, and the policy network is also a 2-hidden-layer MLP with a hidden size of 64. We consider a continuous state space \( \mathcal{S} = [-1, 1]^{50} \) and a discrete action space \( \mathcal{A} = \{0, 1, 2, \ldots, 9\} \). The state distribution $\rho$ is uniform and one-hot feature representation for actions is used.

\begin{figure}[htbp]
    \centering
    \begin{subfigure}{.43\linewidth}
      \centering
      \includegraphics[width=0.9\linewidth]{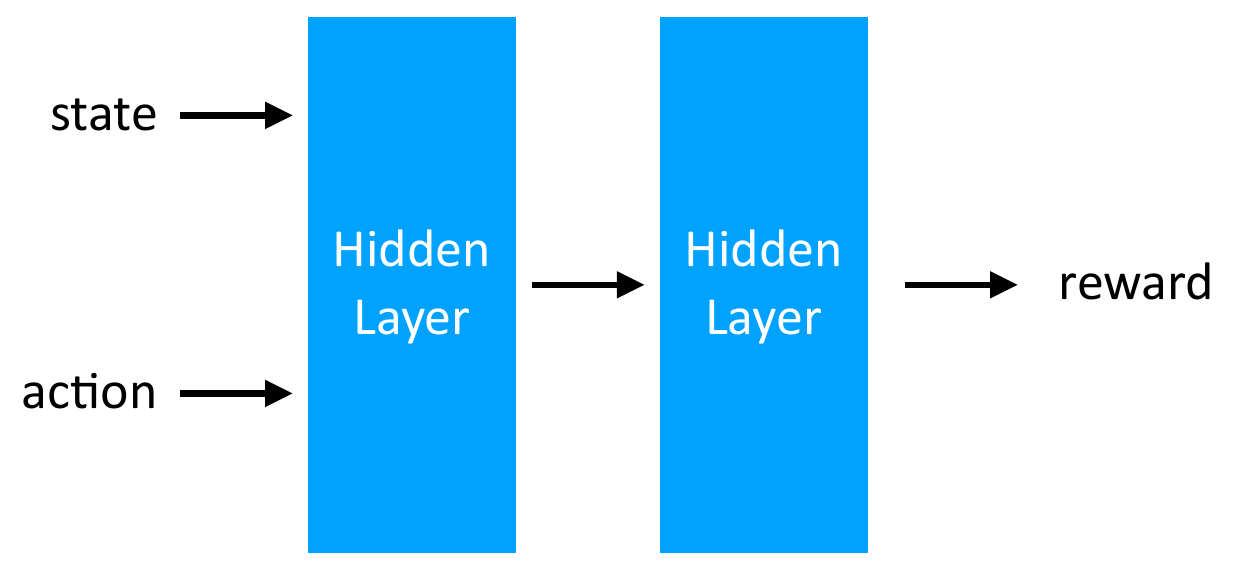}  
      \caption{Reward neural network.}
    \end{subfigure}
    \hfill
    \begin{subfigure}{.49\linewidth}
      \centering
      \includegraphics[width=0.9\linewidth]{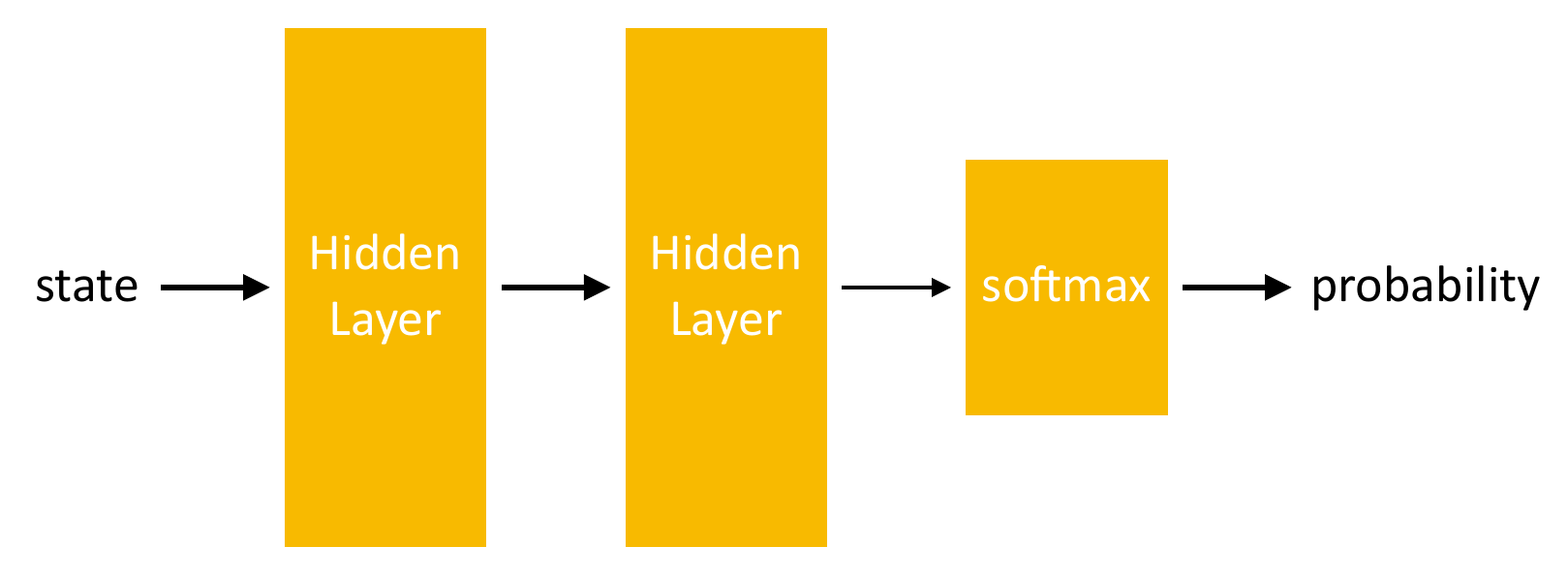} 
      \caption{Policy neural network.}
    \end{subfigure}
    \caption{Architectures of the reward and policy models. }
    \label{fig:neural_network}
\end{figure}

We note that, unlike in the linear bandit case where we could fix the feature representations of the reward and policy models to be the same, in this case, the feature representations of the reward and policy models are purely learned from the given data. The architectures of the reward and policy models are shown in \cref{fig:neural_network}.  All neural networks are optimized using the Adam optimizer \citep{kingma2014adam} with a step size of \(10^{-3}\).

\begin{figure}[htbp]
    \centering
    \includegraphics[width=0.5\linewidth]{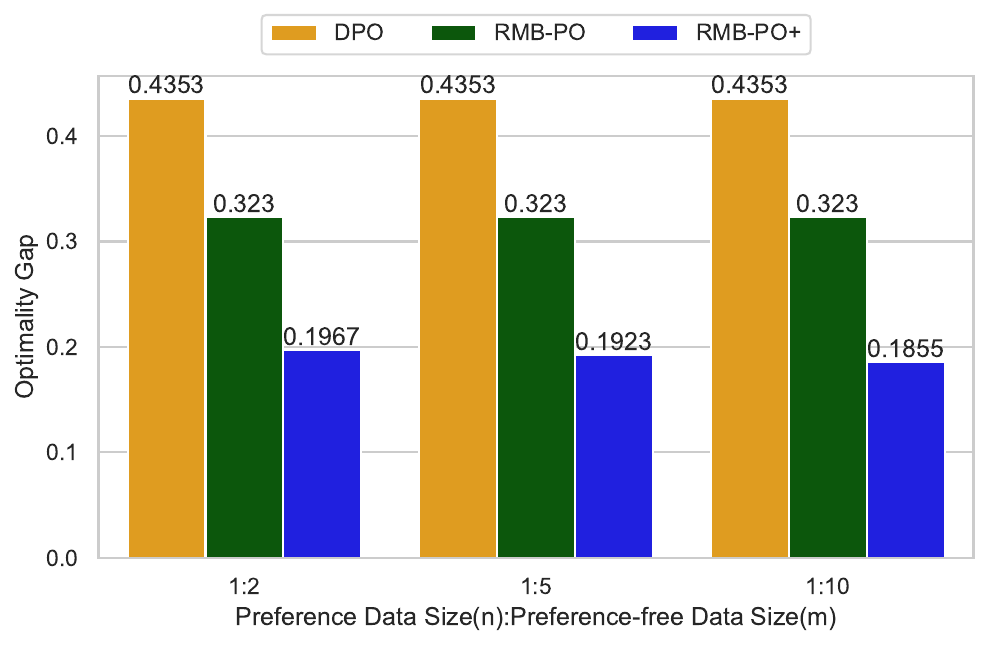}
    \caption{Optimality gap of learned policies in the neural bandit task.}
    \label{fig:neural_bandit}
\end{figure}

We run experiments with varying sizes of preference-free data \(m\) while fixing the preference data size at \(n = 50\). We report the results in \cref{fig:neural_bandit}. First, we observe that RMB-PO and RMB-PO+ significantly outperform DPO. Furthermore, simply using a preference-free data size that is twice as large already improves performance over RMB-PO, and further scaling does not help too much.

\section{Discussion}
\label{sec:discussion}

Our research is related to imitation learning \citep{osa2018survey}, which aims to learn a policy from expert demonstrations. A popular approach to achieve this goal is through behavioral cloning (BC) \citep{Pomerleau91bc}, which trains a policy model by maximizing the likelihood of expert data. Note that the working mechanism of BC is quite similar to DPO, as in \cref{eq:dpo}, where the likelihood of positively preferred actions is increased and that of negatively preferred actions is decreased:
\begin{align*}
    \pi_{\operatorname{BC}} \leftarrow \argmax_{\pi} \sum_{i=1}^{n} \log \pi(a_i|s_i), \quad (s_i, a_i) \sim D_{\operatorname{E}},
\end{align*}
where \( D_{\operatorname{E}} \) is the expert dataset. 

\citet{ghasemipour2019divergence} showed that another class of imitation methods, known as adversarial imitation learning (AIL) methods, (such as GAIL \citep{ho2016gail}), usually performs better than BC. In particular, AIL methods leverage a recovered reward function to perform policy optimization on \dquote{out-of-expert-data} through online interaction, significantly improving performance. Following the formulation in \citep{xu2022understanding}, the training objective of reward-model-based AIL can be re-formulated as 
\begin{align*}
    \pi_{\operatorname{AIL}} \leftarrow \argmin_{\pi} \sum_{s \in \mathcal{S}} \sum_{a \in \mathcal{A}} \labs d_{\pi}(s, a) - \widehat{d}_{\operatorname{E}}(s, a) \rabs ,
\end{align*}
where \( \widehat{d}_{\operatorname{E}} \) is the empirical state-action distribution estimated from \( D_{\operatorname{E}} \), and \( d_{\pi}(s, a) \) is obtained from online interaction. For the optimization objective of AIL, it utilizes states beyond those in the expert dataset (reflected in the summation over all state-action pairs). We notice that \citet{xu2022understanding} theoretically proved that AIL can outperform BC in terms of addressing the distribution shift issue with optimization on \dquote{out-of-expert data}.  The idea of recovering a reward function and using it to perform extensive policy optimization is quite similar to the framework of RLHF.

Additionally, our research is related to transition-model-based reinforcement learning (RL) methods, where the goal is to find an optimal policy through interactions with environments. Many empirical successes suggest that transition-model-based approaches are superior in terms of sample complexity \citep{luo2019algorithmic, janner2019trust}. We do not aim to present a detailed discussion since RL involves lots of concepts and notations. Instead, we would like to highlight that our findings align with the understanding that additional policy optimization on transition-model-generated data is helpful. We would like to refer readers to \citep{hafner2020dream, schrittwieser2020mastering, yu2020mopo, luo2023reward} for the effect of data augmentation in transition-model-based RL methods.

Finally, we note that compared with reward-model-free methods such as DPO \citep{rafailov2023direct}, reward-model-based policy optimization (RMB-PO) methods do not require extra preference annotation. For applications such as language models, training and storing a reward model has been shown to be highly efficient, as demonstrated in \citep{yao2023deepspeed}. The primary challenge in RMB-PO lies in the huge action space during policy optimization. However, this issue can be effectively addressed by computationally efficient methods like those proposed by \citep{dong2023raft,li2023remax}. Notably, \citet{li2023remax} showed that optimizing the language model with prompts-only data can improve performance, a setting that cannot achieved by reward-model-free approaches such as DPO.

\section{Conclusion}

We analyze the errors of policy optimization methods when learning from preferences for alignment. We also conduct experiments to validate our claims. Our results underscore the importance of optimizing policies on out-of-preference data and the power of using a reward model to provide supervision signals.

\bibliography{main}

\begin{thebibliography}{37}
\providecommand{\natexlab}[1]{#1}
\providecommand{\url}[1]{\texttt{#1}}
\expandafter\ifx\csname urlstyle\endcsname\relax
  \providecommand{\doi}[1]{doi: #1}\else
  \providecommand{\doi}{doi: \begingroup \urlstyle{rm}\Url}\fi

\bibitem[Agarwal et~al.(2021)Agarwal, Kakade, Lee, and
  Mahajan]{agarwal2021theory}
Alekh Agarwal, Sham~M Kakade, Jason~D Lee, and Gaurav Mahajan.
\newblock On the theory of policy gradient methods: Optimality, approximation,
  and distribution shift.
\newblock \emph{The Journal of Machine Learning Research}, 22\penalty0
  (1):\penalty0 4431--4506, 2021.

\bibitem[Azar et~al.(2023)Azar, Rowland, Piot, Guo, Calandriello, Valko, and
  Munos]{azar2023general}
Mohammad~Gheshlaghi Azar, Mark Rowland, Bilal Piot, Daniel Guo, Daniele
  Calandriello, Michal Valko, and R{\'e}mi Munos.
\newblock A general theoretical paradigm to understand learning from human
  preferences.
\newblock \emph{arXiv preprint arXiv:2310.12036}, 2023.

\bibitem[Bradley and Terry(1952)]{bradley1952rank}
Ralph~Allan Bradley and Milton~E Terry.
\newblock Rank analysis of incomplete block designs: I. the method of paired
  comparisons.
\newblock \emph{Biometrika}, 39\penalty0 (3/4):\penalty0 324--345, 1952.

\bibitem[Christiano et~al.(2017)Christiano, Leike, Brown, Martic, Legg, and
  Amodei]{christiano2017deep}
Paul~F Christiano, Jan Leike, Tom Brown, Miljan Martic, Shane Legg, and Dario
  Amodei.
\newblock Deep reinforcement learning from human preferences.
\newblock \emph{Advances in Neural Information Processing Systems 30}, pages
  4299--4307, 2017.

\bibitem[Diamond and Boyd(2016)]{diamond2016cvxpy}
Steven Diamond and Stephen Boyd.
\newblock Cvxpy: A python-embedded modeling language for convex optimization.
\newblock \emph{The Journal of Machine Learning Research}, 17\penalty0
  (1):\penalty0 2909--2913, 2016.

\bibitem[Dong et~al.(2023)Dong, Xiong, Goyal, Pan, Diao, Zhang, Shum, and
  Zhang]{dong2023raft}
Hanze Dong, Wei Xiong, Deepanshu Goyal, Rui Pan, Shizhe Diao, Jipeng Zhang,
  Kashun Shum, and Tong Zhang.
\newblock Raft: Reward ranked finetuning for generative foundation model
  alignment.
\newblock \emph{arXiv preprint arXiv:2304.06767}, 2023.

\bibitem[Duchi et~al.(2011)Duchi, Hazan, and Singer]{duchi2011adaptive}
John Duchi, Elad Hazan, and Yoram Singer.
\newblock Adaptive subgradient methods for online learning and stochastic
  optimization.
\newblock \emph{Journal of machine learning research}, 12\penalty0 (7), 2011.

\bibitem[Fishburn et~al.(1979)Fishburn, Fishburn, et~al.]{fishburn1979utility}
Peter~C Fishburn, Peter~C Fishburn, et~al.
\newblock \emph{Utility theory for decision making}.
\newblock Krieger NY, 1979.

\bibitem[Gao et~al.(2023)Gao, Schulman, and Hilton]{gao2023scaling}
Leo Gao, John Schulman, and Jacob Hilton.
\newblock Scaling laws for reward model overoptimization.
\newblock In \emph{Proceedings of the 40th International Conference on Machine
  Learning}, pages 10835--10866, 2023.

\bibitem[Ghasemipour et~al.(2019)Ghasemipour, Zemel, and
  Gu]{ghasemipour2019divergence}
Seyed Kamyar~Seyed Ghasemipour, Richard~S. Zemel, and Shixiang Gu.
\newblock A divergence minimization perspective on imitation learning methods.
\newblock In \emph{Proceedings of the 3rd Conference on Robot Learning}, pages
  1259--1277, 2019.

\bibitem[Hafner et~al.(2020)Hafner, Lillicrap, Ba, and
  Norouzi]{hafner2020dream}
Danijar Hafner, Timothy~P. Lillicrap, Jimmy Ba, and Mohammad Norouzi.
\newblock Dream to control: Learning behaviors by latent imagination.
\newblock In \emph{Proceedings of the 8th International Conference on Learning
  Representations}, 2020.

\bibitem[Ho and Ermon(2016)]{ho2016gail}
Jonathan Ho and Stefano Ermon.
\newblock Generative adversarial imitation learning.
\newblock In \emph{Advances in Neural Information Processing Systems 29}, pages
  4565--4573, 2016.

\bibitem[Janner et~al.(2019)Janner, Fu, Zhang, and Levine]{janner2019trust}
Michael Janner, Justin Fu, Marvin Zhang, and Sergey Levine.
\newblock When to trust your model: Model-based policy optimization.
\newblock In \emph{Advances in neural information processing systems 32}, pages
  12498--12509, 2019.

\bibitem[Kingma and Ba(2015)]{kingma2014adam}
Diederik~P. Kingma and Jimmy Ba.
\newblock Adam: {A} method for stochastic optimization.
\newblock In \emph{Proceedings of the 3rd International Conference on Learning
  Representations}, 2015.

\bibitem[Langford and Zhang(2007)]{langford2007epoch}
John Langford and Tong Zhang.
\newblock The epoch-greedy algorithm for multi-armed bandits with side
  information.
\newblock \emph{Advances in Neural Information Processing Systems 20}, 2007.

\bibitem[Lattimore and Szepesvári(2020)]{banditalgo}
Tor Lattimore and Csaba Szepesvári.
\newblock \emph{Bandit Algorithms}.
\newblock Cambridge University Press, 2020.

\bibitem[Li et~al.(2023)Li, Xu, Zhang, Yu, Sun, and Luo]{li2023remax}
Ziniu Li, Tian Xu, Yushun Zhang, Yang Yu, Ruoyu Sun, and Zhi-Quan Luo.
\newblock Remax: A simple, effective, and efficient method for aligning large
  language models.
\newblock \emph{arXiv preprint arXiv:2310.10505}, 2023.

\bibitem[Lu et~al.(2010)Lu, P{\'a}l, and P{\'a}l]{lu2010contextual}
Tyler Lu, D{\'a}vid P{\'a}l, and Martin P{\'a}l.
\newblock Contextual multi-armed bandits.
\newblock In \emph{Proceedings of the 13th International Conference on
  Artificial Intelligence and Statistics}, pages 485--492, 2010.

\bibitem[Luo et~al.(2023)Luo, Xu, Cao, and Yu]{luo2023reward}
Fan-Ming Luo, Tian Xu, Xingchen Cao, and Yang Yu.
\newblock Reward-consistent dynamics models are strongly generalizable for
  offline reinforcement learning.
\newblock \emph{arXiv preprint arXiv:2310.05422}, 2023.

\bibitem[Luo et~al.(2019)Luo, Xu, Li, Tian, Darrell, and
  Ma]{luo2019algorithmic}
Yuping Luo, Huazhe Xu, Yuanzhi Li, Yuandong Tian, Trevor Darrell, and Tengyu
  Ma.
\newblock Algorithmic framework for model-based deep reinforcement learning
  with theoretical guarantees.
\newblock In \emph{Proceedings of the 7th International Conference on Learning
  Representations}, 2019.

\bibitem[OpenAI(2023)]{openai2023gpt4}
OpenAI.
\newblock Gpt-4 technical report.
\newblock \emph{arXiv preprint arXiv:2303.08774}, 2023.

\bibitem[Osa et~al.(2018)Osa, Pajarinen, Neumann, Bagnell, Abbeel, and
  Peters]{osa2018survey}
Takayuki Osa, Joni Pajarinen, Gerhard Neumann, J.~Andrew Bagnell, Pieter
  Abbeel, and Jan Peters.
\newblock An algorithmic perspective on imitation learning.
\newblock \emph{Foundations and Trends in Robotic}, 7\penalty0 (1-2):\penalty0
  1--179, 2018.

\bibitem[Ouyang et~al.(2022)Ouyang, Wu, Jiang, Almeida, Wainwright, Mishkin,
  Zhang, Agarwal, Slama, Ray, et~al.]{ouyang2022training}
Long Ouyang, Jeffrey Wu, Xu~Jiang, Diogo Almeida, Carroll Wainwright, Pamela
  Mishkin, Chong Zhang, Sandhini Agarwal, Katarina Slama, Alex Ray, et~al.
\newblock Training language models to follow instructions with human feedback.
\newblock \emph{Advances in Neural Information Processing Systems 35}, pages
  27730--27744, 2022.

\bibitem[Pomerleau(1991)]{Pomerleau91bc}
Dean Pomerleau.
\newblock Efficient training of artificial neural networks for autonomous
  navigation.
\newblock \emph{Neural Computation}, 3\penalty0 (1):\penalty0 88--97, 1991.

\bibitem[Rafailov et~al.(2023)Rafailov, Sharma, Mitchell, Ermon, Manning, and
  Finn]{rafailov2023direct}
Rafael Rafailov, Archit Sharma, Eric Mitchell, Stefano Ermon, Christopher~D
  Manning, and Chelsea Finn.
\newblock Direct preference optimization: Your language model is secretly a
  reward model.
\newblock \emph{arXiv preprint arXiv:2305.18290}, 2023.

\bibitem[Russell and Norvig(2010)]{russell2010artificial}
Stuart~J Russell and Peter Norvig.
\newblock \emph{Artificial Intelligence: A Modern Approach}.
\newblock London, 2010.

\bibitem[Schrittwieser et~al.(2020)Schrittwieser, Antonoglou, Hubert, Simonyan,
  Sifre, Schmitt, Guez, Lockhart, Hassabis, Graepel,
  et~al.]{schrittwieser2020mastering}
Julian Schrittwieser, Ioannis Antonoglou, Thomas Hubert, Karen Simonyan,
  Laurent Sifre, Simon Schmitt, Arthur Guez, Edward Lockhart, Demis Hassabis,
  Thore Graepel, et~al.
\newblock Mastering atari, go, chess and shogi by planning with a learned
  model.
\newblock \emph{Nature}, 588\penalty0 (7839):\penalty0 604--609, 2020.

\bibitem[Schulman et~al.(2017)Schulman, Wolski, Dhariwal, Radford, and
  Klimov]{schulman2017ppo}
John Schulman, Filip Wolski, Prafulla Dhariwal, Alec Radford, and Oleg Klimov.
\newblock Proximal policy optimization algorithms.
\newblock \emph{ar{X}iv}, 1707.06347, 2017.

\bibitem[Stiennon et~al.(2020)Stiennon, Ouyang, Wu, Ziegler, Lowe, Voss,
  Radford, Amodei, and Christiano]{stiennon2020learning}
Nisan Stiennon, Long Ouyang, Jeffrey Wu, Daniel Ziegler, Ryan Lowe, Chelsea
  Voss, Alec Radford, Dario Amodei, and Paul~F Christiano.
\newblock Learning to summarize with human feedback.
\newblock \emph{Advances in Neural Information Processing Systems},
  33:\penalty0 3008--3021, 2020.

\bibitem[Sutton and Barto(2018)]{sutton2018reinforcement}
Richard~S Sutton and Andrew~G Barto.
\newblock \emph{Reinforcement {L}earning: {A}n {I}ntroduction}.
\newblock MIT press, 2018.

\bibitem[Touvron et~al.(2023)Touvron, Martin, Stone, Albert, Almahairi, Babaei,
  Bashlykov, Batra, Bhargava, Bhosale, et~al.]{touvron2023llama}
Hugo Touvron, Louis Martin, Kevin Stone, Peter Albert, Amjad Almahairi, Yasmine
  Babaei, Nikolay Bashlykov, Soumya Batra, Prajjwal Bhargava, Shruti Bhosale,
  et~al.
\newblock Llama 2: Open foundation and fine-tuned chat models.
\newblock \emph{arXiv preprint arXiv:2307.09288}, 2023.

\bibitem[Vieillard et~al.(2020)Vieillard, Kozuno, Scherrer, Pietquin, Munos,
  and Geist]{vieillard2020leverage}
Nino Vieillard, Tadashi Kozuno, Bruno Scherrer, Olivier Pietquin, R{\'e}mi
  Munos, and Matthieu Geist.
\newblock Leverage the average: an analysis of kl regularization in
  reinforcement learning.
\newblock \emph{Advances in Neural Information Processing Systems},
  33:\penalty0 12163--12174, 2020.

\bibitem[Xiong et~al.(2023)Xiong, Dong, Ye, Zhong, Jiang, and
  Zhang]{xiong2023gibbs}
Wei Xiong, Hanze Dong, Chenlu Ye, Han Zhong, Nan Jiang, and Tong Zhang.
\newblock Gibbs sampling from human feedback: A provable kl-constrained
  framework for rlhf.
\newblock \emph{arXiv preprint arXiv:2312.11456}, 2023.

\bibitem[Xu et~al.(2022)Xu, Li, Yu, and Luo]{xu2022understanding}
Tian Xu, Ziniu Li, Yang Yu, and Zhi-Quan Luo.
\newblock Understanding adversarial imitation learning in small sample regime:
  A stage-coupled analysis.
\newblock \emph{arXiv preprint arXiv:2208.01899}, 2022.

\bibitem[Yao et~al.(2023)Yao, Aminabadi, Ruwase, Rajbhandari, Wu, Awan, Rasley,
  Zhang, Li, Holmes, et~al.]{yao2023deepspeed}
Zhewei Yao, Reza~Yazdani Aminabadi, Olatunji Ruwase, Samyam Rajbhandari,
  Xiaoxia Wu, Ammar~Ahmad Awan, Jeff Rasley, Minjia Zhang, Conglong Li, Connor
  Holmes, et~al.
\newblock Deepspeed-chat: Easy, fast and affordable rlhf training of
  chatgpt-like models at all scales.
\newblock \emph{arXiv preprint arXiv:2308.01320}, 2023.

\bibitem[Yu et~al.(2020)Yu, Thomas, Yu, Ermon, Zou, Levine, Finn, and
  Ma]{yu2020mopo}
Tianhe Yu, Garrett Thomas, Lantao Yu, Stefano Ermon, James~Y Zou, Sergey
  Levine, Chelsea Finn, and Tengyu Ma.
\newblock Mopo: Model-based offline policy optimization.
\newblock \emph{Advances in Neural Information Processing Systems 33}, pages
  14129--14142, 2020.

\bibitem[Zheng et~al.(2023)Zheng, Chiang, Sheng, Li, Zhuang, Wu, Zhuang, Li,
  Lin, Xing, et~al.]{zheng2023lmsys}
Lianmin Zheng, Wei-Lin Chiang, Ying Sheng, Tianle Li, Siyuan Zhuang, Zhanghao
  Wu, Yonghao Zhuang, Zhuohan Li, Zi~Lin, Eric Xing, et~al.
\newblock Lmsys-chat-1m: A large-scale real-world llm conversation dataset.
\newblock \emph{arXiv preprint arXiv:2309.11998}, 2023.

\end{thebibliography}
\bibliographystyle{plainnat}

\newpage
\appendix

\end{document}